\documentclass[lettersize,journal]{IEEEtran}
\usepackage{amsmath,amsfonts}
 \usepackage{algorithm}
\usepackage{array}
\usepackage[caption=false,font=normalsize,labelfont=sf,textfont=sf]{subfig}
\usepackage{textcomp}
\usepackage{stfloats}
\usepackage{url}
\usepackage{verbatim}
\usepackage{graphicx}
\usepackage{cite}

 \usepackage{algorithmicx,algpseudocode}

\usepackage{amsmath} % assumes amsmath package installed
\usepackage{amssymb}  % assumes amsmath package installed
 \usepackage[ruled, algo2e ]{algorithm2e}

\begin{document}

\title{Bimanual Manipulation of Steady Hand Eye Robots with Adaptive Sclera Force Control: Cooperative vs. Teleoperation Strategies}

\author{ Mojtaba Esfandiari$^{1}$, \IEEEmembership{Graduate Student Member, IEEE}, Peter Gehlbach$^{2}$, {\it Member, IEEE}, \\ Russell H. Taylor$^{3}$, {\it Life Fellow, IEEE}, Iulian I. Iordachita$^{1}$, \IEEEmembership{Senior Member, IEEE}
\thanks{*This work was supported by U.S. National Institutes of Health under the grants numbers 2R01EB023943-04A1, 1R01 EB025883-01A1, R01 EB034397, Research to Prevent Blindness, New York, New York, USA, and gifts by the J. Willard and Alice S. Marriott Foundation, the Gale Trust, Mr. Herb Ehlers, Mr. Bill Wilbur, Mr. and Mrs. Rajandre Shaw, Ms. Helen Nassif, Ms Mary Ellen Keck, Don and Maggie Feiner, Dick and Gretchen Nielsen, and partially by JHU internal funds. }% <-this % stops a space

\thanks{$^{1}$ Mojtaba Esfandiari and Iulian Iordachita are with the Department of Mechanical Engineering and also the Laboratory for Computational Sensing and Robotics at the Johns Hopkins University, Baltimore, MD, 21218, USA (e-mail: {\tt\small mesfand2, iordachita@jhu.edu}).
}%
\thanks{$^{2}$ Peter Gehlbach is with the Wilmer Eye Institute, Johns Hopkins Hospital, Baltimore, MD, 21287, USA. (e-mail: {\tt\small  pgelbach@jhmi.edu})
}%
\thanks{$^{3}$Russell H. Taylor is with the Department of Computer Science and also the Laboratory for Computational Sensing and Robotics at the Johns Hopkins University, Baltimore, MD, 21218, USA. (e-mail: {\tt\small rht@jhu.edu})
}%%
\thanks{Manuscript received August 5, 2024; revised TBD.}}

\markboth{Journal of \LaTeX\ Class Files,~Vol.~14, No.~8, August~2021}%
{Shell \MakeLowercase{\textit{et al.}}: A Sample Article Using IEEEtran.cls for IEEE Journals}

\IEEEpubid{0000--0000/00\$00.00~\copyright~2021 IEEE}

\maketitle

\begin{abstract}
Performing retinal vein cannulation (RVC) as a potential treatment for retinal vein occlusion (RVO) without the assistance of a surgical robotic system is very challenging to do safely. The main limitation is the physiological hand tremor of surgeons. Robot-assisted eye surgery technology may resolve the problems of hand tremors and fatigue and improve the safety and precision of RVC. The Steady-Hand Eye Robot (SHER) is an admittance-based robotic system that can filter out hand tremors and enables ophthalmologists to manipulate a surgical instrument inside the eye cooperatively. However, the admittance-based cooperative control mode does not safely minimize the contact force between the surgical instrument and the sclera to prevent tissue damage.
Additionally, features like haptic feedback or hand motion scaling, which can improve the safety and precision of surgery, require a teleoperation control framework. This work presents a bimanual adaptive teleoperation (BMAT) control framework using SHER 2.0 and SHER 2.1 robotic systems. We integrate them with an adaptive force control (AFC) algorithm to automatically minimize the tool-sclera interaction force. The scleral forces are measured using two fiber Bragg grating (FBG)-based force-sensing tools. We compare the performance of the BMAT mode with a bimanual adaptive cooperative (BMAC) mode in a vessel-following experiment under a surgical microscope. Experimental results demonstrate the effectiveness of the proposed BMAT control framework in performing a safe bimanual telemanipulation of the eye without over-stretching it, even in the absence of registration between the two robots.
\end{abstract}


%%%%%%%%%%%%%%%%%%%%%%%%%%%%%%%%%%%%%%%%%%%%%%%%%%%%%%%%%
\section{INTRODUCTION}

\IEEEPARstart{R}{etinal} vein occlusion (RVO) is the second most prevalent retinal vascular disease, impacting an estimated 28 million individuals globally, according to a meta-analysis study \cite{song2019global}. RVO occurs due to the occlusion of a retinal vein, resulting in eye problems and vision loss. 
There is no established standard for directly treating RVO with a surgical approach. Retinal vein cannulation (RVC) is a proposed solution to treat RVC by delivering a therapeutic agent directly into the affected vein \cite{willekens2017robot}. Freehand RVC is highly challenging due to the micron scale of retinal veins and human hand tremor \cite{singh2002physiological}. Of note, the diameter of the largest retinal veins is on the order of $150 \mu$m \cite{goldenberg2013diameters}, and the root mean square (RMS) amplitude of the hand tremor for an ophthalmic surgeon is reported as $182 \mu$m \cite{riviere2000study}.     

Several surgical robotic systems have been developed to filter out physiological hand tremors and improve the safety of retinal surgery. Some examples of these systems include the Steady Hand Eye Robot (SHER) \cite{mitchell2007development, uneri2010new}, and Preceyes \cite{meenink2013robot},  to name a few  \cite{ida2012microsurgical, nasseri2013introduction, vander2020robotic }.

\begin{figure*}[!t]
    \centering
    \centerline{\includegraphics[width= 1.99 \columnwidth]{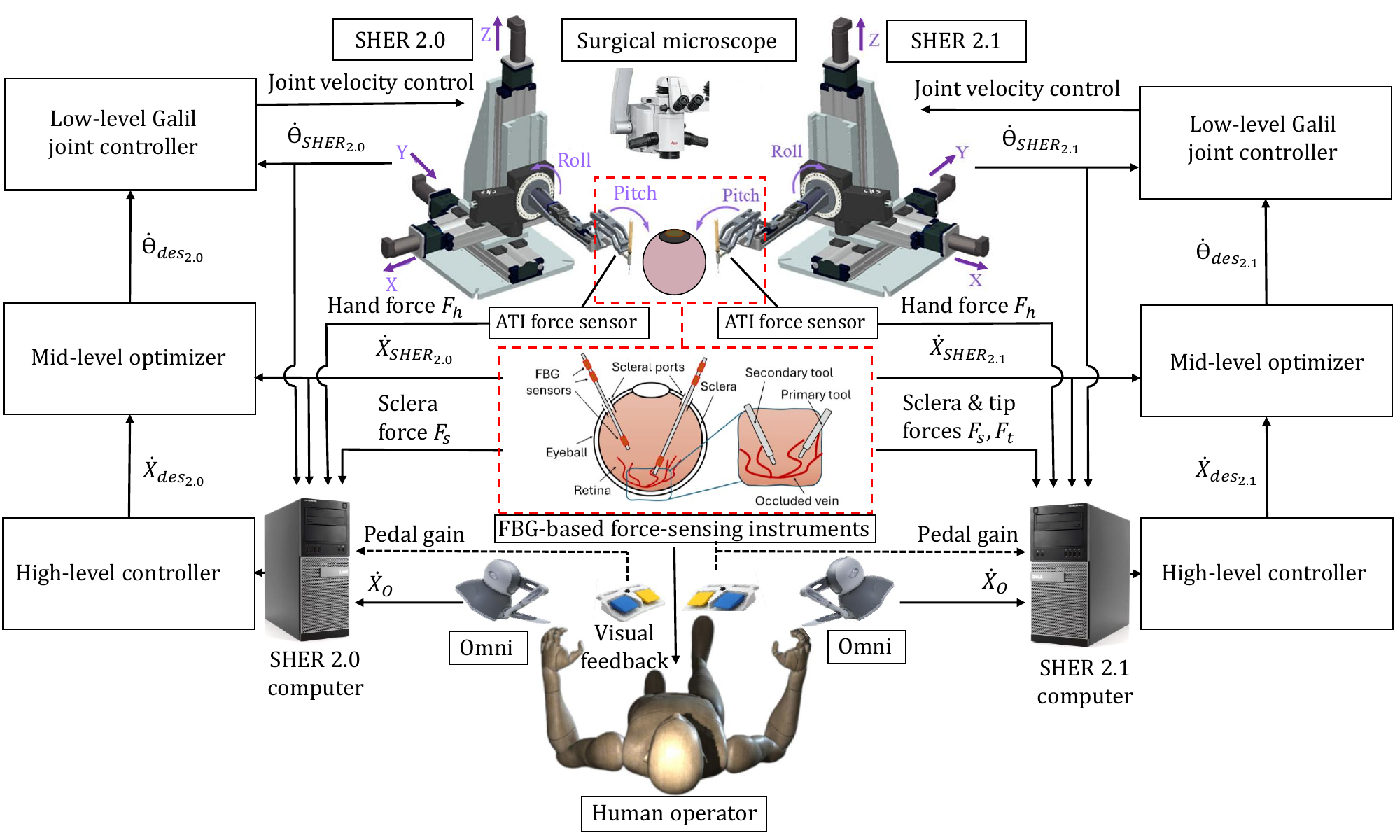}}
    \caption{Bimanual teleoperation architecture with adaptive force control algorithm. The system consists of two Phantom Omni robots, the SHER 2.0 and SHER 2.1, a surgical microscope, two FBG-based force-sensing surgical instruments attached to the SHERs' end-effector to measure the tip force and sclera force, and two foot-pedals to activate and control the SHERs impedance. The figure shows the high-level adaptive force controller, a mid-level optimizer, and a low-level Galil joint velocity controller for SHER 2.0 and SHER 2.1. }
    \label{fig_BiManual_Teleoperation_Framework}
\end{figure*}

The SHER 2.0 and SHER 2.1 are two robot versions developed at Johns Hopkins University. The main control algorithm implemented on SHERs is an admittance-based control where the user and the robot cooperatively manipulate a surgical tool. Using a force/torque sensor attached to the robot handle, the exerted force by the user is measured and converted to a desired end-effector velocity to control the robot \cite{uneri2010new}. An important limitation of this control algorithm is that the force/torque sensor is not sensitive enough to measure the minimal interaction force existing on the surgical instrument at the sclerotomy point, which could result in applying excessive force to the sclera distorting and potentially causing damage. To overcome this issue, Ebrahimi et al. proposed an adaptive sclera force control algorithm \cite{ ebrahimi2021adaptive} using an FBG-based force-sensing tool to automatically minimize the sclera force and keep it within a suggested safe range of 120 mN \cite{ebrahimi2018real}. Notably, FBG sensors have applications in flexible needles or continuum robots, serving purposes such as shape sensing \cite{amirkhani2023design} or force sensing \cite{he2014multi, ryu2020shape}. 
In that study \cite{ebrahimi2021adaptive}, a comparison of the adaptive force control algorithm and the admittance-based control algorithm performance was conducted using a vessel-following experiment in an eye phantom. A primary force-sensing surgical tool was attached to the SHER 2.1 and cooperatively manipulated by the user's dominant hand. Simultaneously, a conventional surgical instrument was directly held by the non-dominant hand, serving as a secondary tool for re-orienting the eye under a surgical microscope \cite{ ebrahimi2021adaptive}. Despite reducing the applied force to the sclera by the dominant hand compared to the admittance-based control mode, controlling the force magnitude on the secondary hand-held tool was impossible, and applying excessively unsafe force to the sclera could not be avoided. Also, this evaluation occurred within a unimanual cooperative control mode without teleoperation capability. Integrating this adaptive force control algorithm in a teleoperation modality has the potential to offer surgeons advanced capabilities, contributing to an improvement in patient safety. Esfandiari et al. implemented this adaptive force control algorithm in a teleoperation modality and compared its performance with a cooperative control mode \cite{esfandiari2023cooperative}. However, these two studies, \cite{ebrahimi2021adaptive, esfandiari2023cooperative}, were conducted in a unimanual robotic framework, which is not a realistic strategy in real retinal surgeries, as they are typically performed bimanually. Doing a bimanual, as opposed to unimanual, robotic manipulation is a challenging problem and needs more safety measures such as task constraints \cite{gottardi2022shared, zhang2023retinal}, robot registration \cite{he2020automatic}, and camera calibration \cite{zhang2023retinal}. These techniques are typically used to dynamically track eye/head movement and satisfy geometrical constraints such as the remote center of motion (RCM) and the relative position of the two robots and the eye, to avoid stretching or compressing the eye. However, the accuracy of these registration-based methods may be adversely affected by eye or head movement, particularly in robotic systems utilizing external optical coherence tomography (OCT) \cite{ahronovich2021review}.   

  To address these limitations, we locally minimize each robot's tool-sclera interaction forces automatically and independently. At the same time, the surgeon bimanually manipulates the eye robots, either using two haptic interfaces in the teleoperation modality or through a hand-over-hand strategy in a cooperative modality. Of note, despite recent advancements in robot intelligence and autonomy, teleoperation is still a dominant method for controlling robots, particularly for performing complex tasks in highly unpredictable situations \cite{peternel2022after}.    
  
  The contributions of this work are as follows: 

  \begin{itemize}
      \item We developed, for the first time in robot-assisted retinal microsurgery research applications, a bimanual adaptive teleoperation (BMAT) control framework integrated with an adaptive sclera force control algorithm on two retinal surgery robotic systems equipped with two FBG-based force-sensing surgical instruments. This framework enables surgeons to safely conduct bimanual teleoperation of the eye, preventing excessive stretch or compression of the sclera, even in the absence of registration of the two robots. 

      \item  We compared the performance of the proposed BMAT control mode with a bimanual adaptive cooperative control mode integrated with the same adaptive force control algorithm. This evaluation occurred within a vessel-following experiment on an eye phantom under a surgical microscope for both sitting and standing postures for each control mode. The objective was to assess and compare the performance of the BMAT and BMAC control modes and also the effect of the user's posture on the performance of the proposed bimanual robotic framework.
      
  \end{itemize}

 Although the sitting posture is preferred in robot-assisted retinal microsurgery, we also study the performance of the proposed BMAT control mode in the standing posture to evaluate its safety for other potential applications in which surgeons may need to (partly) stand during surgery, for example, microvascular anastomosis \cite{harris2022microvascular}.

The remaining sections of this paper are organized as follows: Methods are represented in Section \ref{sec:Methods}. Section \ref{Experimental_setup} describes the experimental setup and procedure. The experimental results are analyzed in Section \ref{sec_experimental_results} and discussed in Section \ref{sec_discussion}. Section \ref{sec:conclusion} concludes the paper.

\begin{figure*}[t!]
    \centering
    \centerline{\includegraphics[width= 1.8 \columnwidth]{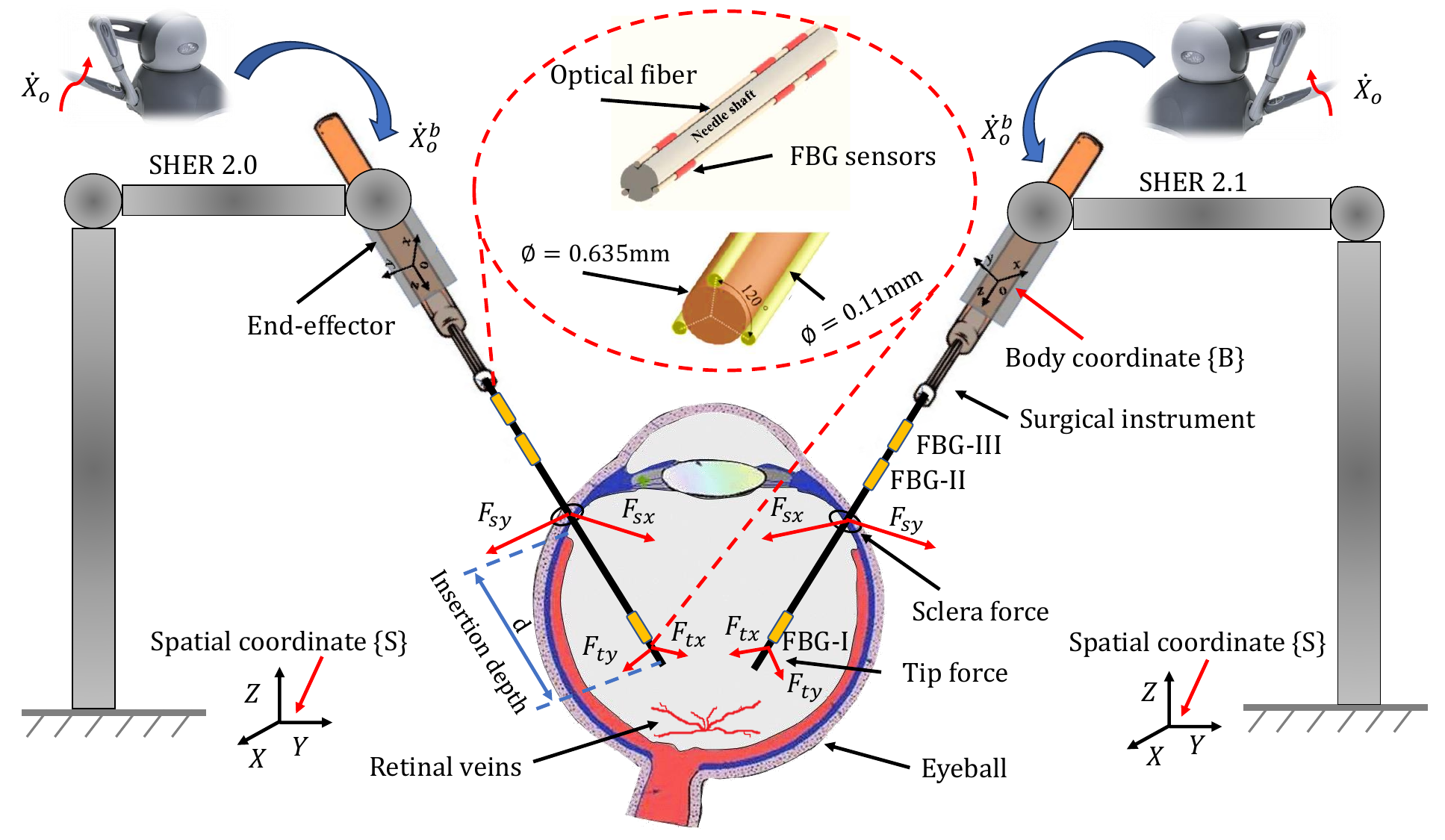}}
    \caption{A diagram of the FBG-based force-sensing surgical instruments connected to the eye robots. These instruments measure the scleral force, tip forces, and insertion depth. The body coordinate, ${B}$, is attached to the robot end-effector, and the spatial coordinate system, ${S}$, is fixed to the robot base. }
    \label{fig:Tool_frames_sclera_force_}
\end{figure*}

%%%%%%%%%%%%%%%%%%%%%%%%%%%%%%%%%%%%%%%%%%%%%%%%%%%%%%%%%%%%%%%%%%%%%%%%
\section{Methods} \label{sec:Methods}

The objective of this study is to extend a previously developed adaptive teleoperation control mode \cite{esfandiari2023cooperative} to two eye robots, the SHER 2.0 and SHER 2.1, and to assess the performance of the two control frameworks: bimanual adaptive cooperative mode and bimanual adaptive teleoperation mode (see Fig. \ref{fig_BiManual_Teleoperation_Framework}), in both sitting and standing postures. In both control modes, the sclera force magnitude is automatically controlled (without the user's commands) by an adaptive force control algorithm to avoid stretching and damaging the sclera.  
Following the suggestion of our clinical lead, we conducted a "vessel-following" experiment, a prototypical task in retinal surgery  \cite{he2019preliminary}. The experimental setup prepared to perform the vessel-following experiment provides a measurement and recording of SHERs' kinematic information, the user's hand force/torque during interaction with the SHERs' handle, and the force applied to the surgical tool at the sclera point and the tooltip.

\subsection{Robot Kinematics}

The Steady-Hand Eye Robot is a 5-degree-of-freedom (5-DoF) robotic manipulator, comprising three translational motions along $XYZ$ cartesian coordinate followed by two rotational, yaw, and pitch degrees of freedom (see Fig. \ref{fig_BiManual_Teleoperation_Framework}). It can carry out a surgical instrument, such as a cannulation tool or a microforceps, at its end-effector. There is no control over the roll motion around the instrument's axis.   

The kinematics of the SHER is described using two coordinate frames: the spatial coordinate $\{S\}$, which remains fixed in space and is situated at the robot's base, and the body coordinate $\{B\}$, which is firmly attached to the robot's handle and moves accordingly (see Fig. \ref{fig:Tool_frames_sclera_force_}) \cite{esfandiari2023cooperative}. 

Utilizing the homogeneous representation $ g_{SB}: \Theta \rightarrow SE(3)$, the SHER end-effector coordinate can be described for any set of joint variables $\Theta \in \mathbb{R}^5$. Here, $g_{SB} \in \mathbb{R}^{4\times 4}$, which constitutes a matrix within the special Euclidean group $SE(3)$, represents the relative configuration of the body coordinate $\{B\}$ to the spatial coordinate $\{S\}$. Using the product of the exponential formula \cite{murray2017mathematical}, the forward kinematics can be represented as follows
\begin{equation} \label{eq_twist1}
g_{SB}(\Theta) = e^{\hat{\xi}_1\theta_1}...e^{\hat{\xi}_5\theta_5}g_{SB}(0),
\end{equation}
where $\theta_i$ represents the $i^{th}$ element of the joint variables $\Theta$, $\hat{\xi}i \in se(3)$ stands for the $i^{th}$ twist, and $g_{SB}(0) \in \mathbb{R}^{4\times 4}$ is the relative configuration of $\{B\}$ to $\{S\}$ when the robot is at home position ($\Theta=\textbf{0}$). 

The twist $\hat{\xi}_i$ is calculated for the three translational joints ($i=1,2,3$) and the two rotational joints ($i=4,5$) using the following formula 
\begin{equation} \label{eq_twist2}
  \hat{\xi}_i = \begin{cases} \quad \begin{bmatrix}
0_{3 \times 3}       & v_i \\
0_{1 \times 3}       & 0  \\
\end{bmatrix}, \quad \quad \quad \quad \textrm{for} \quad  i = 1,2,3 \vspace{0.2cm} \\ 
\quad \begin{bmatrix}
\hat{\omega}_i       & -\omega_i \times q_i \\
0_{1 \times 3}      & 0  \\
\end{bmatrix}, \quad \textrm{for} \quad i = 4,5
\end{cases}
\end{equation}
where $v_i$ is a unit vector representing the $i^{th}$ prismatic joint and is along with its positive direction, $\omega_i$ is a unit vector representing the $i^{th}$ revolute joint and is directed to the positive rotation axis (counterclockwise) of this joint, and $q_i$ denotes an arbitrary point on this rotation axis. Both $v_i$ and $\omega_i$ are represented in the $\{S\}$ coordinate when the robot is at home position.

Substituting the twist \eqref{eq_twist2} into \eqref{eq_twist1}, it is possible to develop a mapping between the robot joint velocity vector $\dot{\Theta} \in \mathbb{R}^5$ and the robot end-effector velocity vector $V_{SB}^b \in \mathbb{R}^6$. Here, the superscript \( b \) indicates that the velocity of the body coordinate \( \{B\} \) relative to the spatial coordinate \( \{S\} \) is expressed in \( \{B\} \). This relationship is given by:
\begin{equation} \label{eqtwist4}
V_{SB}^b = J_{SB}(\Theta)\Dot{\Theta}
\end{equation}
where \( J_{SB} \in \mathbb{R}^{6 \times 5} \) is the robot Jacobian and is formulated as:
\begin{align}\label{eqtwist5}
J_{SB}(\Theta) &=
\begin{bmatrix}
\xi_1^\dagger & \dots & \xi_5^\dagger 
\end{bmatrix},\\
\xi_i^\dagger &= Ad^{-1}_{e^{\hat{\xi}_i\theta_i} \cdots e^{\hat{\xi}_5\theta_5}g_{SB}(0)}\xi_i \nonumber
\end{align}
where \( Ad_{g}: \mathbb{R}^6 \rightarrow \mathbb{R}^6 \) represents the adjoint transformation of the rigid body transformation \( g = (p, R) \in SE(3) \), and is defined as follows \cite{murray2017mathematical}:
\begin{equation}
Ad_{g} = \begin{bmatrix}
R & \hat{p}R \\ 
0 & R 
\end{bmatrix}_{6\times 6}.
\label{eq: adjoint_transformation}
\end{equation}

\begin{figure*}[t!]
    \centering
    \includegraphics[scale=0.37]{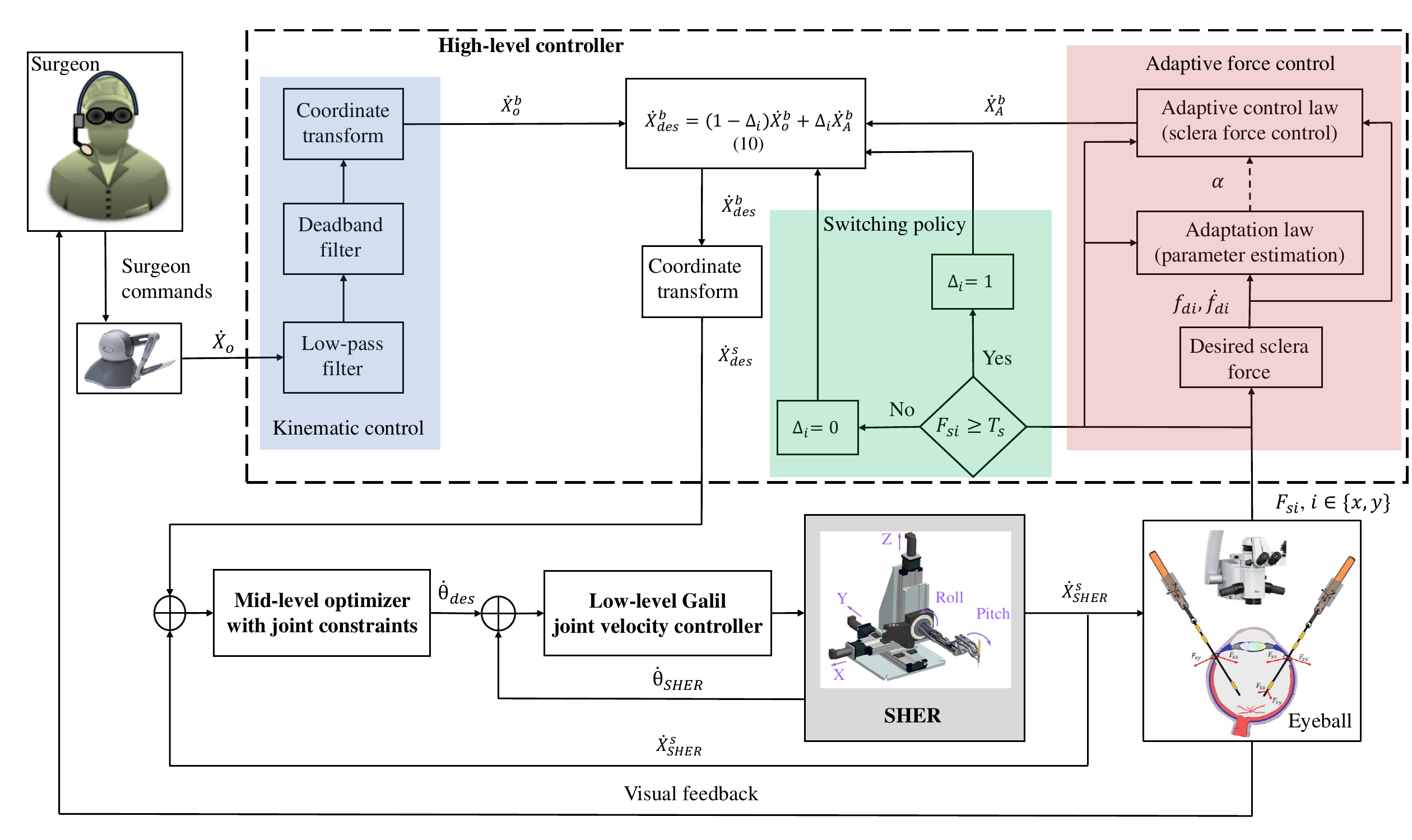}
    \caption{ Control diagram of the bimanual teleoperation modality integrated with adaptive sclera force control algorithm.  }
    \label{fig:Bimanual_Tele_control_structure}
\end{figure*}

\subsection{Adaptive Force Control of the SHER}
This adaptive force control algorithm is automatically activated if each of the measured sclera force components, $F_{sx}$ and $F_{sy}$ (see Fig. \ref{fig:Tool_frames_sclera_force_}), surpasses a specified safety threshold, $T_s$. Then, the robot is directed along a safe sclera force trajectory \cite{ebrahimi2021adaptive} minimizing the tool-sclera interaction force. The desired sclera forces, $f_{dx}$ and $f_{dy}$, are defined as exponentially decaying trajectories as follows
\begin{equation}
\label{x-component of the desired force trajectory}
    \begin{aligned}
    f_{dx}=\frac{T_{s}sign(F_{sx})}{2}(e^{-(t-t_{x})}+1),
    \end{aligned}
\end{equation}
\begin{equation}
\label{y-component of the desired force trajectory}
    \begin{aligned}
    f_{dy}=\frac{T_{s}sign(F_{sy})}{2}(e^{-(t-t_{y})}+1),
    \end{aligned}
\end{equation}
where $t_{i}$ represents the time at which $F_{si}$ surpasses the safe threshold.

The desired linear velocities of the SHER's end-effector along the $x$ and $y$ directions in the body frame $\{B\}$ can be determined by the adaptive force controller by estimating the stiffness of the sclera tissue, denoted by $\alpha_{i}$ $(i=x$ or $y)$, using the following equations
\begin{equation}
\label{AFC desired linear velocity}
    \begin{aligned}
    \dot{X}^{b}_{A_{i}}(t)=\alpha_{i}\dot{f}_{di}(t)-K_{fi}\Delta{f}_{i}(t),\quad i=\{x,y\}, 
    \end{aligned}
\end{equation}
\begin{equation}
\label{AFC tissue stiffness}
    \begin{aligned}
    \dot{\alpha}_{i}=-\Gamma_{i}\dot{f}_{di}(t)\Delta{f}_{i}(t),\quad i=\{x,y\}
    \end{aligned}
\end{equation}
in which $\Delta{f}_{i}=F_{si}-f_{di}$ denotes the sclera force error, and $\Gamma_{i}$ and $K_{fi}$ represent constant gains for the adaptation law and the force tracking error, respectively. $\alpha_{i}$ stands for the estimated tissue compliance, and $\dot{f}_{di}$ is the derivative of $f_{di}$. This adaptive sclera force control algorithm applies to both cooperative and teleoperation control modalities. In the cooperative control mode, the remaining components of the SHER end-effector velocity are generated by the admittance-based controller. In contrast, in the teleoperation control mode, they are derived from the PHANTOM Omni's end-effector velocity $\dot{X}_o$. The proposed BMAT control mode, integrated with an adaptive sclera force control algorithm, consists of three primary components: a high-level controller, a mid-level optimizer with joint constraints, and a low-level joint velocity controller (Galil), see Fig. \ref{fig:Bimanual_Tele_control_structure}.
% \vspace{5mm}

The high-level controller consists of two primary components: Firstly, there is a kinematic controller that translates the end-effector velocity of the Omni, denoted as $\dot{X}_o$, into the SHER's body coordinate as $\dot{X}_o^b$ (see Fig. \ref{fig:Tool_frames_sclera_force_}). It is then pre-multiplied by $K = diag(\kappa_1, \kappa_2, \kappa_3, \kappa_4, \kappa_5, \kappa_6)$, a user-defined motion scaling matrix, to provide an intuitive motion on the SHER end-effector velocity, i.e., $\dot{X}_{des}^b= K \dot{X}_o^b$. Secondly, an adaptive sclera force controller that receives inputs from the sclera force components $F_{sx}$ and $F_{sy}$, which are measured in the body coordinate $\{B\}$ by the force-sensing instrument. This controller generates the desired adaptive end-effector velocity in the SHER body coordinate, denoted as $\dot{X}_A^b$ in \eqref{AFC desired linear velocity}.

% \vspace{-10mm}
This hybrid kinematic-force control method determines the desired SHER velocity by switching between $\dot{X}_A^b$ and $\dot{X}_o^b$ based on the magnitude of the sclera force. When the scleral force components fall below a safe threshold $T_s$, the SHER is controlled based on kinematic teleoperation using the Omni velocity $\dot{X}_o^b$; otherwise, the adaptive force controller takes charge, enforcing the SHER's end-effector velocity through $\dot{X}_A^b$ to minimize scleral force components (see Fig. \ref{fig:Bimanual_Tele_control_structure}). As a result, the desired end-effector velocity of SHER, $\dot{X}_{des}^b$, is generated based on the following hybrid kinematic-force control algorithm: 
\begin{align}
&\dot{X}_{des}^b = \nonumber\\ 
 &\begin{bmatrix}
    (1 - \Delta_x)\kappa_1 & 0 & \textbf{0}_{1 \times 4} \\ 
    0 & (1 - \Delta_y)\kappa_2 & \textbf{0}_{1 \times 4} \\ 
    \textbf{0}_{4 \times 2} & & diag(\kappa_3, \kappa_4, \kappa_5, \kappa_6)
\end{bmatrix} \dot{X}_o^b \nonumber \\ 
+ &\begin{bmatrix}
    diag(\Delta_x, \Delta_y) & \textbf{0}_{2\times 4} \\ 
    \textbf{0}_{4 \times 2} & \textbf{0}_{4 \times 4}
\end{bmatrix}\begin{bmatrix}
    \dot{X}_{A_x}^b \\
    \dot{X}_{A_y}^b \\ 
    \textbf{0}_{4 \times 1}
\end{bmatrix},
    \label{eq_hybrid_force-velocity_control}
\end{align}
% \vspace{0mm}
where $\Delta_i$ ($i= x$ or $y$) is a binary variable automatically set by the control switching policy (Algorithm \ref{alg_hyb_kin_force_switch_policy}) based on the magnitude of sclera force components $F_{si}$ ($i= x$ or $y$). $\Delta_i=1$ means the adaptive force control algorithm for the corresponding sclera force component ($x$ or $y$) is activated. Otherwise, when $\Delta_i=0$, the adaptive force control algorithm gets deactivated, and Equation \eqref{eq_hybrid_force-velocity_control} becomes equivalent to the kinematic control $\dot{X}_{des}^b= K \dot{X}_o^b$. 
\begin{algorithm}
\caption{Hybrid Kinematic-Force Control Switching Policy}\label{alg_hyb_kin_force_switch_policy}
\KwIn{Sclera force components $F_{si}$, $i=x,y$.}
\KwOut{$\Delta_i$ and $\dot{X}_{A_i}^b$, $i=x,y$.}
$\Delta_i \gets 0$;\\
\If{$F_{si} \geq T_s$}{
$t_i \gets t$; \Comment{Save current time $t$}\\ 
$\Delta_i \gets 1$; \Comment{Activate adaptive force control}\\
  \While{\ $|F_{si}| > 0.75 T_s$}{
    $\Delta_i \gets 1$; \Comment{Keep adaptive force control activated}\\
    $f_{di} \gets$ \eqref{x-component of the desired force trajectory}, \eqref{y-component of the desired force trajectory}; \Comment{Calculate desired force} \\ 
    $\dot{X}_{A_i}^b \gets$ \eqref{AFC desired linear velocity}, \eqref{AFC tissue stiffness}; \Comment{Calculate adaptive velocities}  
  }
  $\Delta_i \gets 0$; \Comment{Switch to kinematic control for axis $i$}
}
\end{algorithm}

 The mid-level optimizer computes the optimal desired joint velocities, denoted as $\dot{\Theta}_{des}$, for the SHER while accounting for its joint limit constraints. Finally, the low-level Galil joint velocity controller generates the necessary control commands to reach the desired joint velocities. Other signals include $\dot{\Theta}_{SHER}$ (actual joint velocities of the SHER), $\dot{X}_{SHER}^s$ (actual end-effector velocity in the spatial coordinate of the SHER), and $F_{t}$ (tooltip force) (see Fig. \ref{fig:Bimanual_Tele_control_structure}).

\vspace{-1pt}
%%%%%%%%%%%%%%%%%%%%%%%%%%%%%%%%%%%%%%%%%%%%%%%%%%%%%%%%%%%
\section{Experiments} \label{Experimental_setup}

This experiment aims to evaluate the performance of the proposed BMAT control mode regarding patient safety in terms of minimizing the tool-sclera interaction forces and avoiding the stretch and compression of the sclera in the absence of registration between the two robots. Also, the performance of the BMAT framework is compared with that of a bimanual adaptive cooperative framework in both sitting and standing postures. 

\begin{figure*}[t!]
    \centering
    \includegraphics[scale=0.48]{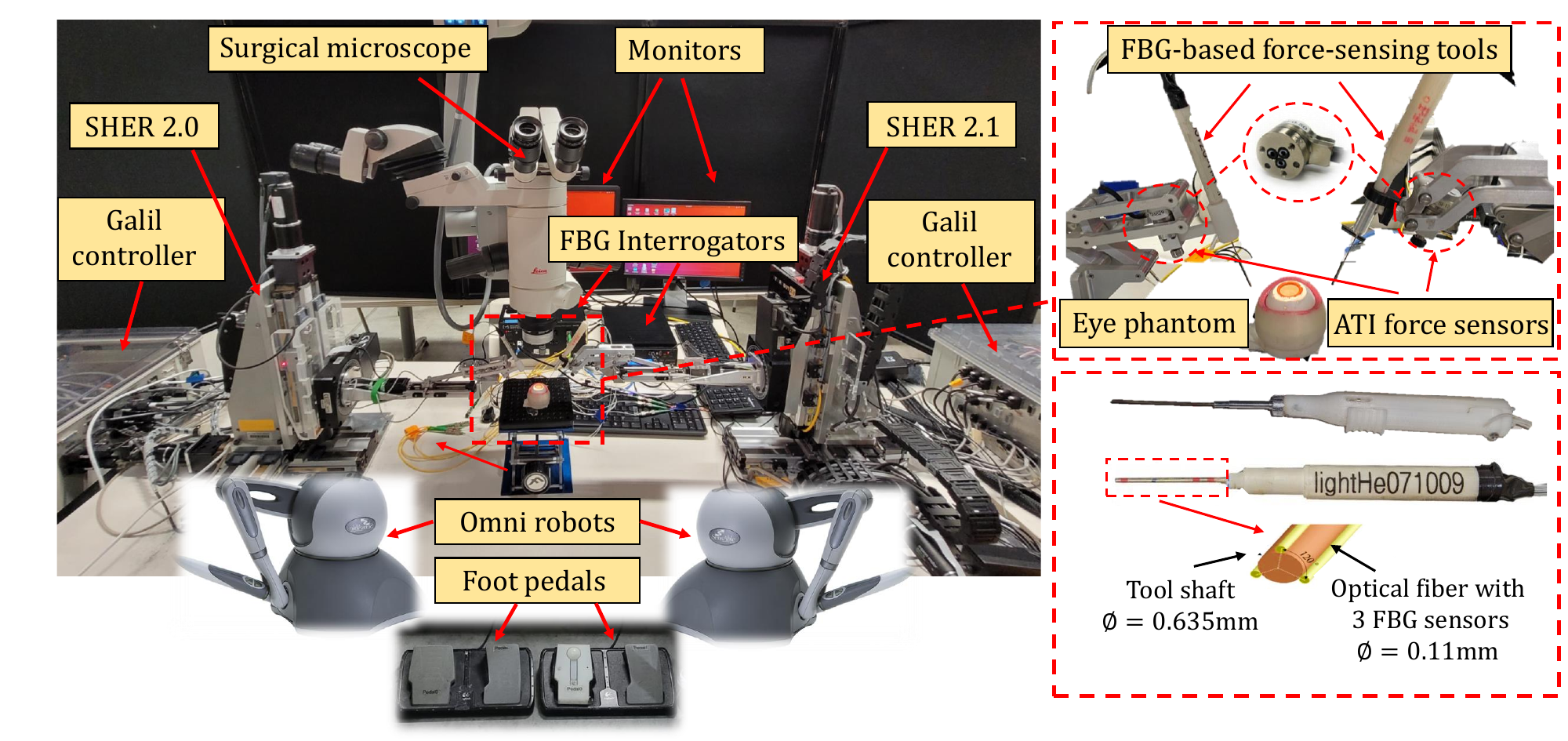}
    \caption{The experiment setup includes the SHER 2.1 and the SHER 2.0, two PHANTOM Omni haptic interfaces, two FBG interrogators, a surgical microscope, and an armrest (left); an eye phantom and two ATI force/torque sensors (top right); two FBG-based force-sensing tools attached to the SHERs' handles (bottom right). }
    \label{fig_Experiment_setup_BMac_BMAT_standing_sitting_Mojtaba}
\end{figure*}

\begin{figure*}[t!]
    \centering
    \includegraphics[scale=0.47]{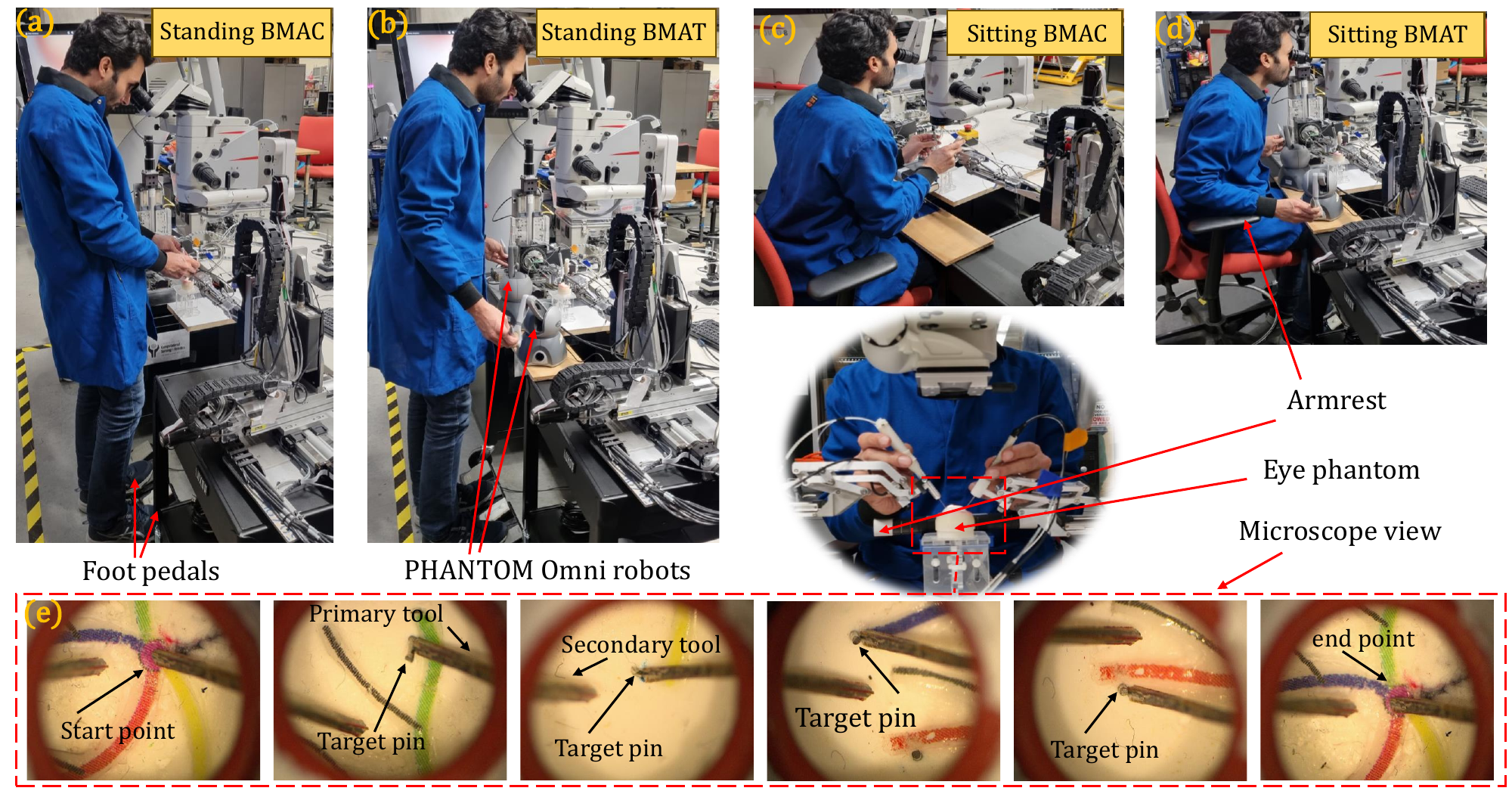}
    \caption{Vessel-following experiment procedure for four combinations of control modes and user posture including (a) standing BMAC, (b) standing BMAT, (c) sitting BMAC, and (d) sitting BMAT. (e) The user activates each robot using a foot pedal and follows a random order of colored vessels by the dominant (right) hand, touches a target pin at the end of each vessel by the tooltip of the dominant hand, and re-orients the eye phantom using the non-dominant hand under a surgical microscope.} 
    \label{fig_Experiment_procedure_BMAC_BMAT_sitting_standing}
\end{figure*}

% The following sections briefly describe our experimental setup and procedure as well as the considered performance metrics for the patient safety and surgeon comfort.
\subsection{Experimental Setup}
 The setup components comprised the SHER 2.1 on the right holding the primary force-sensing tool, the SHER 2.0 on the left holding the secondary force-sensing tool, two robot motion controllers (Galil motion controllers, CA, USA), and two PHANTOM Omni robots (SensAble Technologies Inc., MA, USA). The two PHANTOM Omnis serve as a haptic interface for the bimanual teleoperation control of SHER 2.0 and SHER 2.1. Two 6-DoF force/torque sensors (Nano17, ATI Industrial Automation, NC, USA) are integrated into the SHERs' end-effector to measure the user hand's force and torque exerted on the robot handle and to control the robot in the cooperative mode with an admittance-based control algorithm \cite{uneri2010new}. Two FBG-based force-sensing surgical instruments (tools) measure the scleral force applied to the eye phantom. The tool shaft has three FBG fibers arranged at 120-degree intervals, housing a total of nine FBG sensors (Technica Optical Components, China) distributed along the three fibers \cite{he2019dual} (see Fig. \ref{fig_Experiment_setup_BMac_BMAT_standing_sitting_Mojtaba}). Optical signals from these FBG sensors are received by two FBG interrogators, (HYPERION SI155, Luna) for the primary force-sensing tool and (SM130, Micron Optics, Inc.) for the secondary fore-sensing tool, enabling the estimation of the tip force, sclera force, and insertion depth by measuring the wavelength shifts of the nine FBG sensors attached to each tool. The force read at the sclera point has two components, $F_{sx}$ and $F_{sy}$, perpendicular to the tool shaft and are used in the adaptive sclera force control algorithm to always maintain the sclera force level below a safe threshold (see Fig. \ref{fig:Tool_frames_sclera_force_}). These force-sensing surgical instruments are calibrated using a method outlined by \cite{he2014multi}. An eye phantom is designed to simulate the retinal veins of the human eye. An armrest supports the user's hand during the vessel-following experiment under a surgical microscope (Leica Microsystems) (see Fig.\ref{fig_Experiment_setup_BMac_BMAT_standing_sitting_Mojtaba}). All these devices are interconnected to the two robot computers through a TCP-IP connection.

\subsection{Experimental Procedure}

To assess the performance of the BMAC and BMAT control modes, a user, who is an expert in bimanual manipulation of the SHERs, is asked to conduct a vessel-following experiment in sitting and standing postures for each control mode (see Fig. \ref{fig_Experiment_procedure_BMAC_BMAT_sitting_standing}a-d). The task involves tracking a colored trajectory on the retinal surface of an eye phantom that simulates retinal veins. The colored trajectory comprises four different colors: red, green, blue, and yellow. A pin is attached at the end of each colored vessel to specify a target point, and the user is asked to touch the pins with the tooltip. This requires pinpoint accuracy in tool use by the user. The user starts the experiment by approaching the tip of the primary tool, manipulated by the dominant hand, to the start point at the intersection of the four colored vessels, tracking the four colored vessels in a random order, and touching the target pins at the end of each colored vessel. Once all colors are tracked, the user returns to the intersection point (see Fig. \ref{fig_Experiment_procedure_BMAC_BMAT_sitting_standing}e). This procedure is considered one trial. More specific details of the procedure are outlined in \cite{he2019preliminary}. The experiment includes 5 sets of tests collected on five different days. For each test, 10 trials are repeated following a random order of colored vessels, resulting in 50 trials for each mode: standing BMAC, sitting BMAC, standing BMAT, and sitting BMAT. All trials are conducted under a surgical microscope with an illuminator to optimize visualization.

In the bimanual cooperative control mode, the user holds the tool handle and manipulates the eye robots in a hand-over-hand strategy (Figs. \ref{fig_Experiment_procedure_BMAC_BMAT_sitting_standing}a, \ref{fig_Experiment_procedure_BMAC_BMAT_sitting_standing}c). In contrast, in the bimanual teleoperation mode, the user indirectly interacts with the SHERs using the PHANTOM Omni interfaces. Two foot-pedals activate each robot separately. The more the foot pedal is pushed, the less impedance is observed in the SHER handle and the surgical instrument.

\begin{figure*}[t!]
	\centering
    \subfloat[]{
    \centering
    \includegraphics[width=0.31 \textwidth]{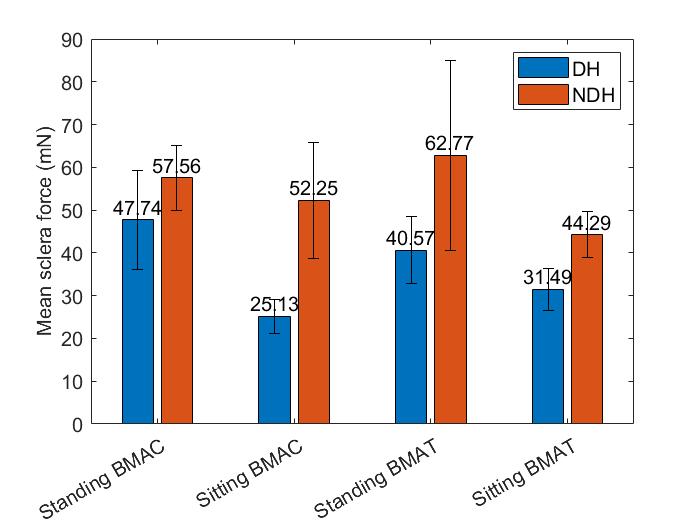}
    \label{fig_BarChart_MeanSclera}}  
    \subfloat[]{
    \centering
    \includegraphics[width=0.31 \textwidth]{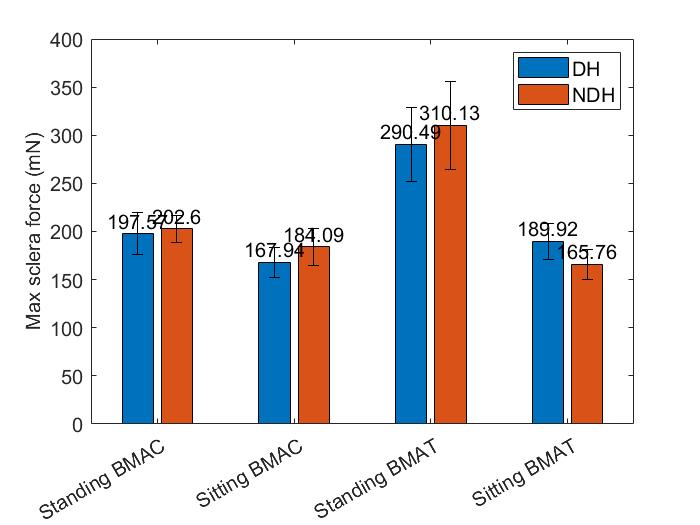}
    \label{fig_BarChart_MaxSclera}}
    \subfloat[]{
    \centering
    \includegraphics[width=0.31 \textwidth]{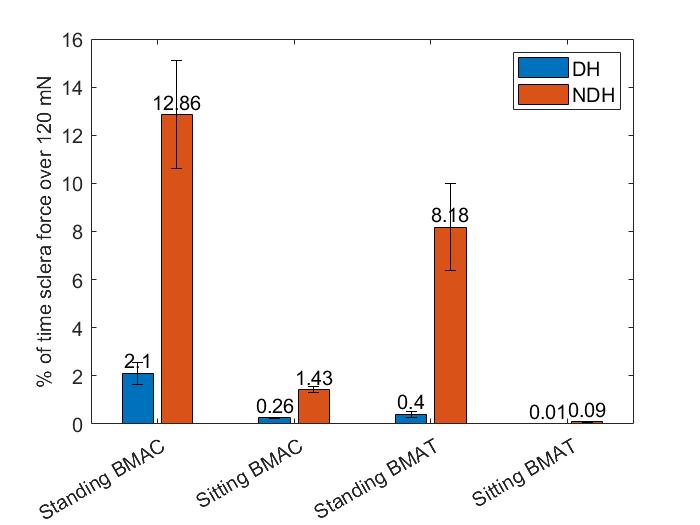}
    \label{fig_BarChart_forceOver120}}
	\caption{ Comparison of the mean sclera force (a), maximum sclera force (b), and percent of time when the sclera force is over 120 mN (c) for the BMAC and BMAT control modes in sitting and standing postures for the dominant hand (DH) and non-dominant hand (NDH).  }
\label{fig_BarChart}
\end{figure*}

\begin{figure*}[t!]
	\centering
    \subfloat[DH]{
    \centering
    \includegraphics[width=0.7\textwidth]{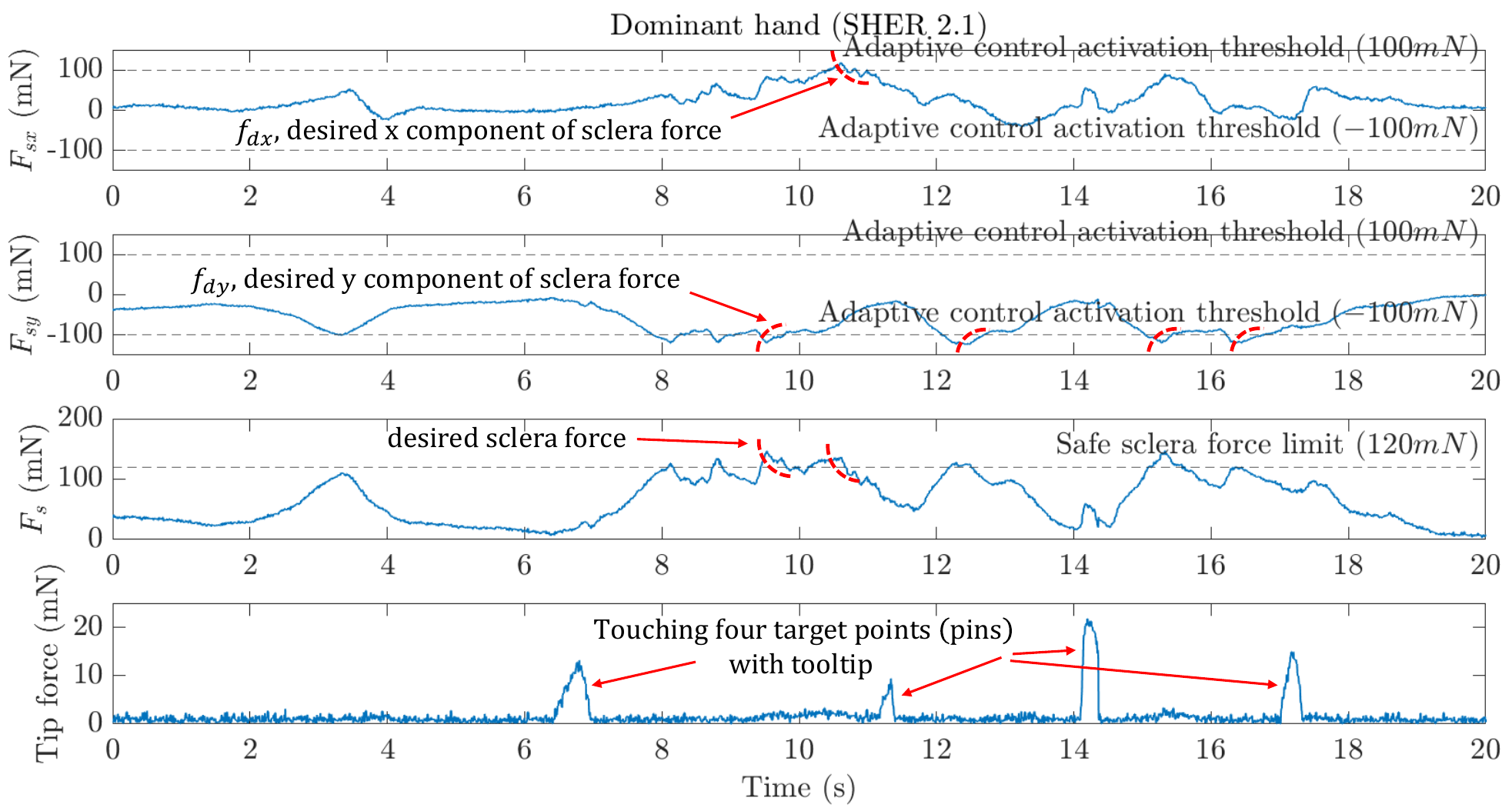}
    \label{fig_Example_Sclera_force_BMAT_mojtaba_sitting_dominantHand}} \\ 
    \subfloat[NDH]{
    \centering
    \includegraphics[width=0.7\textwidth]{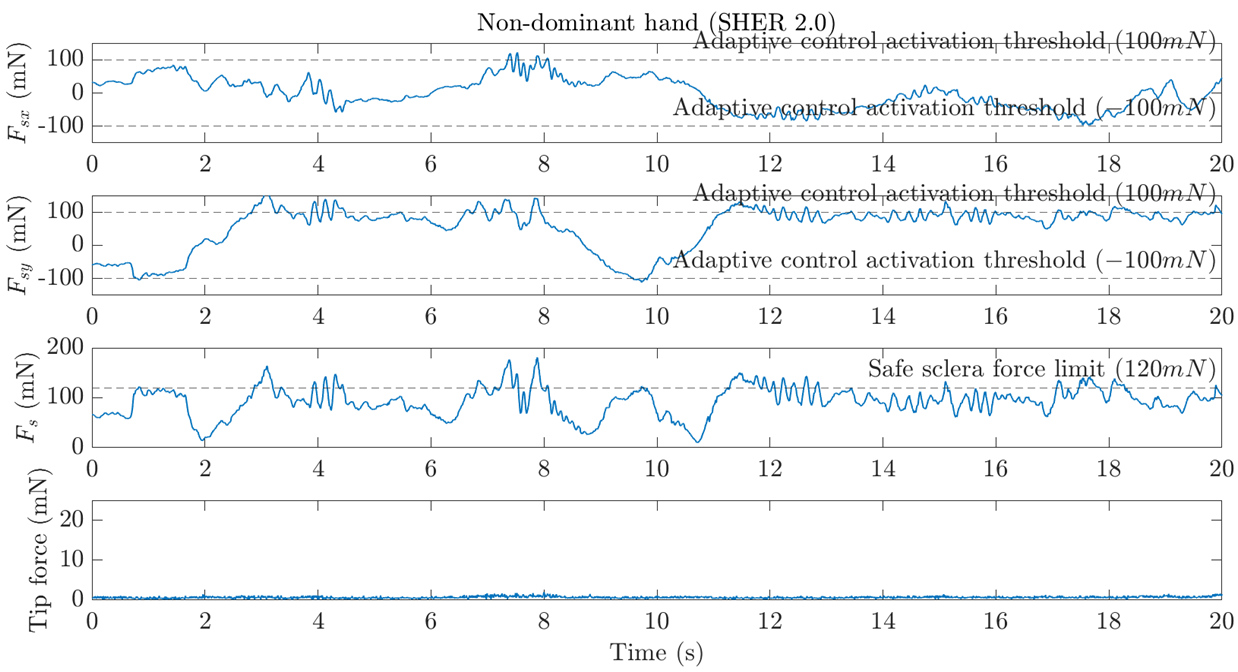}
    \label{fig_Example_Sclera_force_BMAT_mojtaba_sitting_NondominantHand}}
	\caption{An example of the performance of the adaptive force control algorithm on limiting the sclera force components $F_{sx}$ and $F_{sy}$ and sclera force norm $F_s$ for the dominant hand (a) and the non-dominant hand (b) during sitting bimanual adaptive teleoperation control mode. Upon reaching the 100 mN threshold, the adaptive sclera force control algorithm is triggered to minimize the sclera force components, namely $F_{sx}$ and $F_{sy}$, along a desired exponentially decaying force trajectory (dashed curves). The objective of the adaptive force controller is to keep the norm of sclera force $F_s$ below a safe threshold of 120 mN \cite{ebrahimi2018real}. The four peaks in the tip force for the dominant hand show that the four target pins at the end of the colored vessels are touched by the tooltip. }
\label{fig_Example_Sclera_force_BMAT_mojtaba_sitting}
\end{figure*}

During the sitting BMAC and sitting BMAT, the user's hand is stabilized by an armrest to further reduce the hand tremor (see Figs. \ref{fig_Experiment_procedure_BMAC_BMAT_sitting_standing}c, \ref{fig_Experiment_procedure_BMAC_BMAT_sitting_standing}d), whereas this armrest is lacking in the standing postures.  Regardless of the control mode, the user uses the dominant hand to manipulate the primary force-sensing tool and the non-dominant hand to manipulate a secondary tool to reorient the eye phantom to reach optimum visualization under the microscope. Of note, bimanual manipulation is a standard method of performing retinal surgery.

All kinematic and force information measured from the robots, the ATI force sensors, and the FBG-based force-sensing tools for each vessel-following trial are recorded in separate .csv files and subsequently analyzed using MATLAB. The statistical analyses were performed using a paired t-test, MATLAB's \texttt{ttest2()} function, assuming unequal variance (Welch's t-test). Statistical significance is inferred when p$<$0.05.

\section{Experimental Results} \label{sec_experimental_results}
 To our knowledge, implementing a bimanual adaptive teleoperation control mode integrated with an adaptive sclera force control algorithm on a microsurgical robotic system for retinal microsurgery applications is unprecedented. 
 The performance of the proposed BMAT mode is compared with that of a BMAC mode in sitting and standing postures for the user by considering several important metrics related to patient safety and surgeons' comfort. For example, some factors having to do with patient safety, including the mean and maximum sclera force and the percentage of aggregate completion times during which the recorded sclera force exceeded the safety threshold of 120 mN, are reported for both dominant and non-dominant hand for all four control modes (see Fig. \ref{fig_BarChart}). Other factors related to surgeons' intuition and comfort are also reported such as human-robot interaction force/torque and task completion time (Table \ref{table:mean_max_force_Bilateral_coop_tele_standing_sitting}). Completion time is also important, determining the general operational time and cost. These factors are analyzed and reported for a single expert user and may not be extrapolated to non-expert users who are not at their learning curve plateau \cite{zhao2023human}.

\begin{figure}[b!]
	\centering
    \subfloat[]{
    \centering
    \includegraphics[width=0.51 \textwidth]{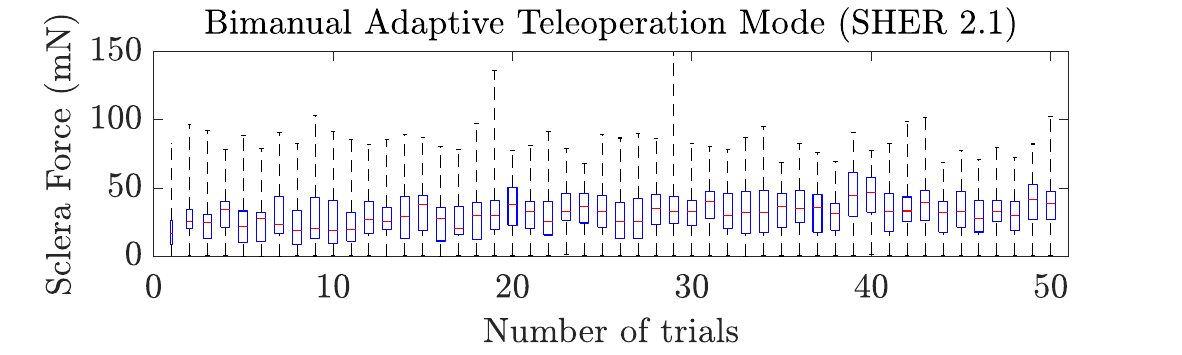}
    \label{fig_scleraForce_BMAT_right_Mojtaba}} \\ 
    \subfloat[]{
    \centering
    \includegraphics[width=0.51 \textwidth]{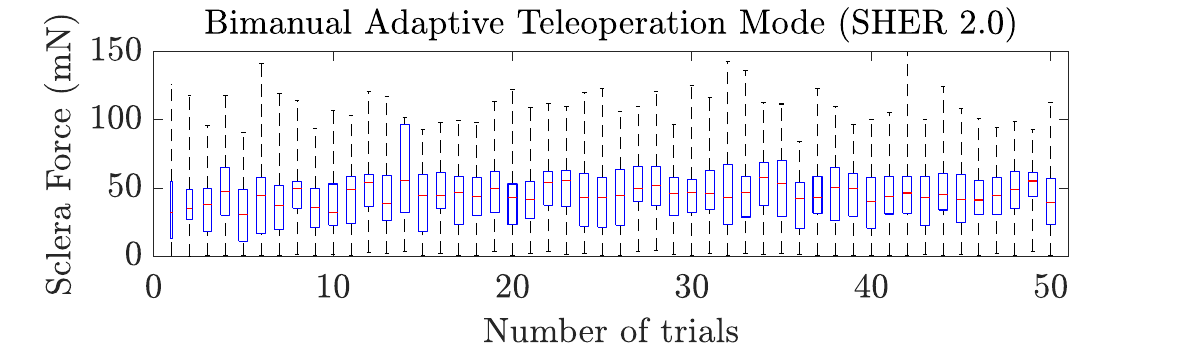}
    \label{fig_scleraForce_BMAT_left_Mojtaba}}
	\caption{The mean of the tool-sclera interaction forces is calculated over 50 trials for the sitting BMAT mode for (a) dominant hand and (b) non-dominant hand. This figure shows that the user is an expert in using the robots in both control modes. }
\label{fig_scleraForce_BMAT_sitting_Mojtaba}
\end{figure}

\begin{table*}[t!]
\centering
    \caption{Mean and maximum sclera force, handle force/torque, mean completion time, and the percentage of time when the sclera force is over 120 mN. The results are reported for the BMAC and the BMAT control modes, for both sitting and standing postures, and for both the dominant hand (right robot, SHER 2.1) and non-dominant hand (left robot, SHER 2.0), during the vessel-following experiment.  }
    \includegraphics[width= 1.62 \columnwidth]{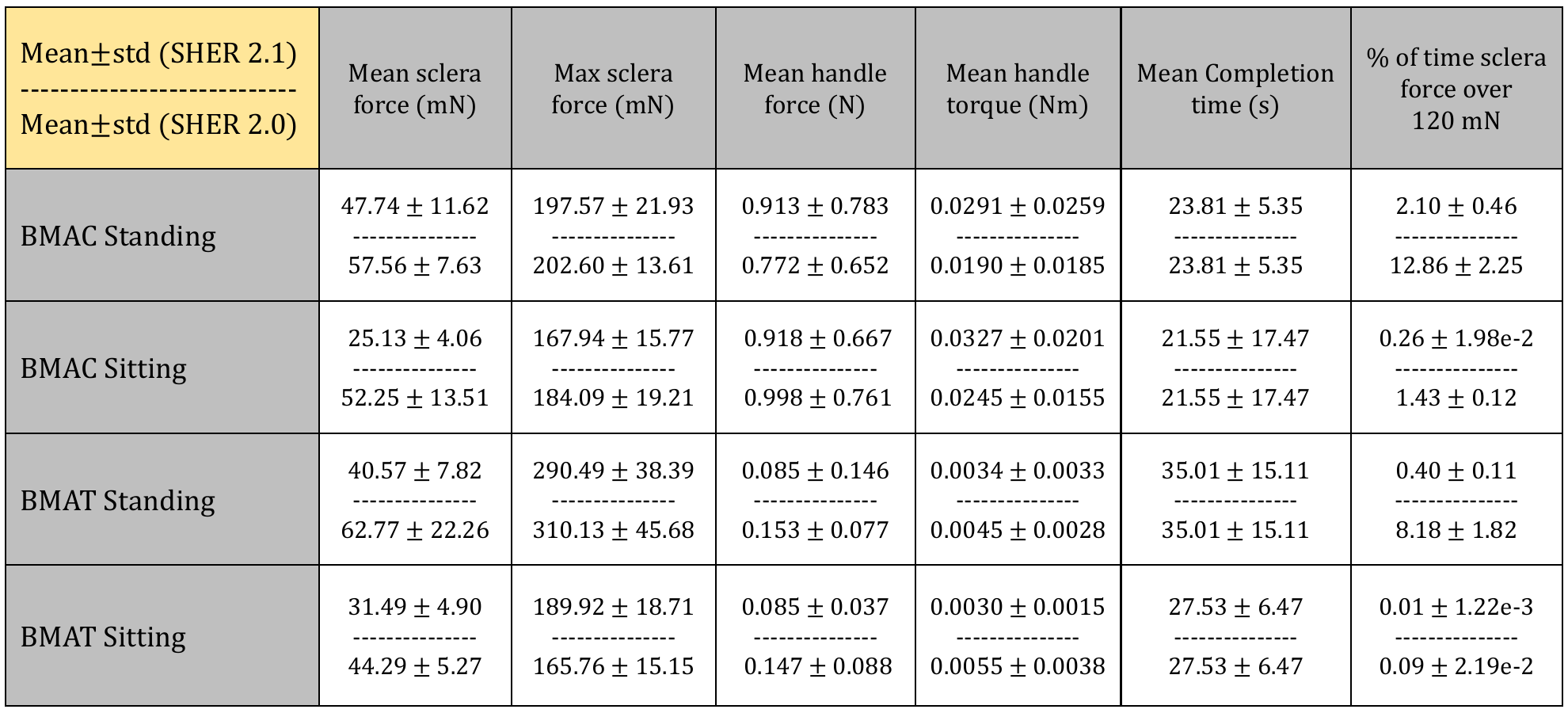}
\label{table:mean_max_force_Bilateral_coop_tele_standing_sitting}
\end{table*}

\begin{table}[b!]
\fontsize{8pt}{8pt}\selectfont
\caption{The p-values for standing BMAC vs. sitting BMAC, for mean and maximum sclera force, mean completion time, and the percentage of time during which the sclera force is over 120 mN, for both the dominant hand (DH) and the non-dominant hand (NDH). }
\label{table:pvalue_BMAC_sitting_BMAC_standing_mojtaba}
\begin{tabular}{ |p{1.0cm}||p{1.33cm}|p{1.33cm}|p{1.33cm}|p{1.33cm}|  }
 \hline
 \multicolumn{5}{|c|}{Standing BMAC vs. Sitting BMAC} \\
 \hline
 p-value& Mean sclera force & Max sclera force& Mean completion time & $\%$ of time force over 120 mN \\
 \hline
 DH   & 0  & 1.3695e-11 &   0.3854
 & 0
 \\
 \hline
 NDH &   0.0179 & 2.8404e-07   & 0.3854 & 0
\\
 \hline
 \end{tabular}
\end{table}

\begin{table}[b!]
\fontsize{8pt}{8pt}\selectfont
\caption{The p-values for standing BMAT vs. sitting BMAT. }
\label{table:pvalue_BMAT_sitting_BMAT_standing_mojtaba}
\begin{tabular}{ |p{1.0cm}||p{1.33cm}|p{1.33cm}|p{1.33cm}|p{1.33cm}|  }
 \hline
 \multicolumn{5}{|c|}{Standing BMAT vs. Sitting BMAT} \\
 \hline
 p-value& Mean sclera force & Max sclera force& Mean completion time & $\%$ of time force over 120 mN \\
 \hline
 DH   
 &   7.6327e-10  & 
 0 & 5.2077e-04 & 0
 \\
 \hline
 NDH &  4.7921e-07  &  0  & 5.2077e-04
 & 0
\\ \hline
 \end{tabular}
\end{table}

\begin{table}[b!]
\fontsize{8pt}{8pt}\selectfont
\caption{The p-values for sitting BMAC vs. sitting BMAT.}
\label{table:pvalue_BMAC_sitting_BMAT_sitting_mojtaba}
\begin{tabular}{ |p{1.0cm}||p{1.33cm}|p{1.33cm}|p{1.33cm}|p{1.33cm}|  }
 \hline
 \multicolumn{5}{|c|}{Sitting BMAC vs. Sitting BMAT} \\
 \hline
 p-value& Mean sclera force & Max sclera force& Mean completion time & $\%$ of time force over 120 mN \\
 \hline
 DH   
 &  2.6583e-10   & 7.2431e-09
  & 0.0267 & 0
 \\
 \hline
 NDH &  2.4950e-04  &  7.8348e-07  & 0.0267
 & 0
\\ \hline
 \end{tabular}
\end{table}

 The user, in both sitting and standing postures, is supposed to follow a random order of colored vessel trajectory while approaching the tooltip as close as possible to the bottom of the eye phantom without touching it and guide the tooltip toward a target pin attached to the end of each vessel and slightly touch it by the tooltip. 

 \subsection{Sclera Force Control Performance}

 The adaptive sclera force control algorithm dynamically adjusts the SHER 2.0 and SHER 2.1 end-effector velocity to maintain the sclera force below a safe threshold of 120 mN. This feature lets the surgeon safely manipulate the eye without applying unsafe force to the sclera. Using this technique, it is possible to safely perform bimanual robot manipulation of the eye without registering the two robots yet avoid stretching the eye during bimanual manipulation. Figure \ref{fig_Example_Sclera_force_BMAT_mojtaba_sitting} describes an example of the functionality of the sclera force control algorithm during the sitting BMAT control mode for both dominant and non-dominant hands. It is observed that once the sclera force $F_{s}$ reaches the 100 mN threshold, the adaptive sclera force control algorithm gets activated and tries to minimize the sclera force components, $F_{sx}$ and $F_{sy}$, along a desired exponentially decaying force trajectory. The proposed safe limit for sclera force is 120 mN \cite{ebrahimi2018real}. The 100 mN threshold is selected for activating the adaptive force control to provide a 20 mN offset below the safe limit to give the AFC algorithm enough time to reduce the scleral force before it reaches 120 mN. The presence of four peaks in the tip force data for the dominant hand (Fig. \ref{fig_Example_Sclera_force_BMAT_mojtaba_sitting_dominantHand}) indicates that the tooltip made contact with the four target points (pins) located at the ends of the colored vessels. Also, an example of the mean sclera force distribution for all 50 trials for the sitting BMAT control mode is provided in Fig. \ref{fig_scleraForce_BMAT_sitting_Mojtaba}, which indicates that the user is an expert in robot use and showed a consistent performance in this mode for both dominant and non-dominant hands. Similar performance is observed for the other three control/posture modes.

Here, the standing and sitting postures for each of the two BMAC and BMAT control modes are compared. Moreover, the sitting BMAC and the sitting BMAT are compared as well (p-values are reported in Tables \ref{table:pvalue_BMAC_sitting_BMAC_standing_mojtaba} -- \ref{table:pvalue_BMAC_sitting_BMAT_sitting_mojtaba}), to analyze the feasibility of BMAC and BMAT and examine the effects of user's posture on their performance.  

\subsection{Standing BMAC vs. Sitting BMAC}
For example, by comparing the standing BMAC and sitting BMAC, it is observed that the user applied a significantly smaller maximum force of 167.94 $\pm$ 15.77 mN by his dominant hand during the sitting BMAC compared to this value in the standing BMAC which is 197.57 $\pm$ 21.93 mN ($p$ $<$ 0.01) (see Fig. \ref{fig_BarChart_MaxSclera} and Tables \ref{table:mean_max_force_Bilateral_coop_tele_standing_sitting}, \ref{table:pvalue_BMAC_sitting_BMAC_standing_mojtaba}). This difference is also significant for the non-dominant hand. The percentage of aggregate time when the sclera force is above 120 mN for the dominant hand is 0.26 $\pm$ 0.01 for the sitting BMAC and 2.10 $\pm$ 0.46 for the standing BMAC ($p$ $<$ 0.01). There is no significant difference in task completion time between the standing and sitting BMAC ($p$ $>$ 0.05).

\subsection{Standing BMAT vs. Sitting BMAT}
Similarly, by comparing the sitting BMAT and standing BMAT, it is observed that the user demonstrated better performance sitting. For example, the mean sclera force for the dominant hand during sitting BMAT is 31.49 $\pm$ 4.90 mN, whereas this value for the standing BMAT is 40.57 $\pm$ 7.82 mN ($p$ $<$ 0.01) (see Fig. \ref{fig_BarChart_MeanSclera} and Tables \ref{table:mean_max_force_Bilateral_coop_tele_standing_sitting} and \ref{table:pvalue_BMAT_sitting_BMAT_standing_mojtaba}), or the maximum sclera force for the dominant hand in the standing BMAT is 290.49 $\pm$ 38.89 mN while this value is significantly reduced to 189.92 $\pm$ 18.71 mN for the sitting BMAT ($p$ $<$ 0.01). Similar improvement is also observed for the mean completion time and the percentage of aggregate time when sclera force is above 120 mN. Based on these findings, we conclude that the user outperforms in the sitting posture because the arms are stabilized using an armrest, reducing hand tremors and fatigue. Furthermore, it is difficult to handle both foot pedals in standing mode as the user is supposed to hold their weight on their heels and pivot to push each pedal separately. Sometimes, the user did not handle the pedals efficiently, especially the left one, related to the non-dominant hand. This resulted in higher stiffness of the robots and, therefore, higher maximum sclera force for the non-dominant hand. Perhaps, using a single pedal instead of two to activate both robots may resolve this issue in the standing posture. Although the sitting posture allows a superior performance that makes it a preferred option for robot-assisted retinal microsurgery, the standing posture also allows acceptable and safe performance, which could potentially be used for other surgeries during which the surgeons may decide to partly stand during the operation such as microvascular anastomosis \cite{harris2022microvascular}. 

\subsection{Sitting BMAC vs. Sitting BMAT}

Finally, by comparing the results of the BMAT and the BMAC modes in the sitting posture (Tables \ref{table:mean_max_force_Bilateral_coop_tele_standing_sitting}, \ref{table:pvalue_BMAC_sitting_BMAT_sitting_mojtaba}), it is observed that bimanual adaptive teleoperation showed comparable or better performance, in some terms, to the bimanual adaptive cooperative mode. For example, the mean sclera force for the non-dominant hand for the sitting BMAT is 44.29 $\pm$ 5.27 mN, which is less than this value for the sitting BMAC (52.25 $\pm$ 13.51, $p$ $<$ 0.01). Also, the percentage of aggregate time with a scleral force higher than 120 mN is significantly less for both dominant and non-dominant hands for the sitting BMAT than the sitting BMAC ($p$ $<$ 0.01) (see Fig. \ref{fig_BarChart_forceOver120}). This demonstrates lower scleral forces with a sitting BMAT. This could be, in part, due to the slower end-effector velocity during the teleoperation mode compared to the cooperative mode, which in turn results in a longer task completion time for the sitting BMAT mode (27.53 $\pm$ 6.47 s) compared to the sitting BMAC mode (21.55 $\pm$ 17.47 s, $p$ $<$ 0.05). The maximum sclera force for the non-dominant hand for the sitting BMAT is less than that of the sitting BMAC (165.76 $\pm$ 15.15 mN vs. 184.09 $\pm$ 19.21 mN, $p$ $<$ 0.01), whereas this is the other way around for the dominant hand (189.92 $\pm$ 18.71 mN vs. 167.94 $\pm$ 15.77 mN, $p$ $<$ 0.01). Based on our understanding, this has to do with the repositioning capability of the teleoperation mode. In the cooperative mode, the user can finish the task by holding the pedals during the entire vessel-following trial, whereas, during the teleoperation mode, the motion scaling capability necessitates, mainly for the dominant hand, several repeated unclutching, repositioning of the Omni handles, and clutching again to follow a trajectory. This results in a higher stiffness at the SHERs' handle and the surgical instrument at the beginning of each clutching, which resulted in a higher mean and maximum sclera force for the dominant hand for the BMAT mode; however, it is still in a safe range and on the same level of magnitude as the BMAC mode.

 %%%%%%%%%%%%%%%%%%%%%%%%%%%%%%%%%%%%%%%%%%%%%%%%%%%%%%%%%%
 \section{Discussion and Future Work} \label{sec_discussion}

We proposed a bimanual adaptive teleoperation framework equipped with adaptive force control capability using FBG-based force-sensing instruments in this work.

As opposed to other bimanual teleoperation frameworks that typically depend on the robot registration to satisfy the RCM constraints \cite{zhang2023retinal}, our proposed adaptive teleoperation framework is capable of performing a safe bimanual manipulation of the human eye despite the absence of registration of the two robots.  
Both surgical tools, including the secondary tool used for rotating the eyeball, are force-sensing tools that provide sclera force measurements for the adaptive force control algorithm to adjust scleral forces automatically to maintain them below a safe threshold.

 Based on the observed results, the proposed BMAT control mode demonstrated the desired performance during the vessel-following experiment. It could be a promising technology with new capabilities to improve surgeons' comfort, dexterity, and patients' safety.

One limitation of this work is that the experiments are conducted by a single user. This is a pilot study to analyze the performance of the different control modes. Moreover, since the force-sensing tools are fragile and expensive to build, it is important to work with expert users with enough training in bimanual teleoperation control of the robots. 
For future work, the effectiveness of the proposed BMAT control mode should be further examined using other types of biological models, such as chicken embryo or pig eye. Moreover, integrating this BMAT mode with haptic force feedback \cite{fu2018framework, ji2022improving, sadeghnejad2023using, sadeghnejad2023virtual}, augmented \cite{brizzi2017effects} or mixed reality \cite{li2024bilateral} technology, and learning from demonstration (LfD) \cite{10585785 ,baumkircher2021performance} may improve non-experts' ability to efficiently manipulate and control the robots \cite{sun2020new} and improve the safety of surgery.

%%%%%%%%%%%%%%%%%%%%%%%%%%%%%%%%%%%%%%%%%%%%%%%%%%%%%%%%%%%%%
\section{Conclusion} \label{sec:conclusion}

This work performed a vessel-following experiment on a bimanual adaptive teleoperation framework with an adaptive sclera force control algorithm on SHER 2.0 and SHER 2.1 using two FBG-based force-sensing tools. The tools were equipped with FBG sensors capable of measuring the force at the tool sclerotomy and the tooltip. The performance of this bimanual teleoperation mode was compared with a bimanual cooperative mode utilizing a single expert user by doing a vessel-following experiment inside an eye phantom for two sitting and standing postures. Using this force control technique, it is possible to safely perform bimanual robotic manipulation of the eye without registering the two robots. Also, a comparison between the BMAT and the BMAC modes in the sitting posture demonstrates that the BMAT mode shows comparable performance to the BMAC mode and can potentially improve the safety and precision of surgery by providing surgeons with additional capabilities, including motion scaling and repositioning.

% \section*{Acknowledgments}
% This should be a simple paragraph before the References to thank those individuals and institutions who have supported your work on this article.

\bibliographystyle{IEEEtran}
\bibliography{biblio}

\newpage

\vspace{11pt}

\vfill

\end{document}